\documentclass[10pt,a4paper]{article}

\usepackage{amssymb}
\usepackage{graphicx}
\usepackage{authblk}

\usepackage{url}

\begin{document}

% \mainmatter  % start of an individual contribution

\title{\LARGE \bf The CARESSES EU-Japan project: \\
  making assistive robots culturally competent%
\thanks{Paper presented at: \textit{Ambient Assisted Living, Italian
  Forum}. Genova, Italy, June 12--15, 2017.}%
}

\author[1]{Barbara Bruno}
\author[2]{Nak Young Chong}
\author[3]{Hiroko Kamide}
\author[4]{Sanjeev Kanoria}
\author[5]{Jaeryoung Lee}
\author[2]{Yuto Lim}
\author[6]{Amit Kumar Pandey}
\author[7]{Chris Papadopoulos}
\author[8]{Irena Papadopoulos}
\author[9]{Federico Pecora}
\author[9]{Alessandro Saffiotti}
\author[1]{Antonio Sgorbissa}

\affil[1]{University of Genova, Via Opera Pia 13, 16145 Genova, Italy}
\affil[2]{Japan Advanced Institute of Science and Technology, 1-1
Asahidai, Nomi, Ishikawa 923-1292, Japan}
\affil[3]{Nagoya University, Furocho, Chikusaku, Nagoya, Aichi 464-8601, Japan}
\affil[4]{Advinia Health Care Limited LTD, Regents Park Road 314, London N3 2JX, United Kingdom}
\affil[5]{Chubu University, 1200 Matsumoto-cho, Kasugai, Aichi 487-8501, Japan}
\affil[6]{Softbank Robotics Europe SAS, Rue Colonel Pierre Avia 43, 75015 Paris, France.}
\affil[7]{University of Bedfordshire, Park Square, Luton LU1 3JU, United Kingdom}
\affil[8]{Middlesex University Higher Education Corporation, The Burroughs, Hendon, London NW4 4BT, United Kingdom}
\affil[9]{\"Orebro University, Fakultetsgatan 1, S-70182 \"Orebro, Sweden}

\date{June 5, 2017}
	
%\author{Alfred Hofmann%
%\thanks{Please note that the LNCS Editorial assumes that all authors have used
%the western naming convention, with given names preceding surnames. This determines
%the structure of the names in the running heads and the author index.}%
%\and Ursula Barth\and Ingrid Haas\and Frank Holzwarth\and\\
%Anna Kramer\and Leonie Kunz\and Christine Rei\ss\and\\
%Nicole Sator\and Erika Siebert-Cole\and Peter Stra\ss er}
%
% \authorrunning{Bruno et al.}

%\institute{Springer-Verlag, Computer Science Editorial,\\
%Tiergartenstr. 17, 69121 Heidelberg, Germany\\
%\url{http://www.springer.com/lncs}}


%\toctitle{Lecture Notes in Computer Science}
% \tocauthor{Bruno et al.}
\maketitle

\begin{abstract}
The nursing literature shows that cultural competence is an important
requirement for effective healthcare.  We claim that personal assistive
robots should likewise be culturally competent, that is, they should be
aware of general cultural characteristics and of the different forms
they take in different individuals, and take these into account while
perceiving, reasoning, and acting.  The CARESSES project is an
Europe-Japan collaborative effort that aims at designing, developing and
evaluating culturally competent assistive robots. These robots will be
able to adapt the way they behave, speak and interact to the cultural
identity of the person they assist.  This paper describes the approach
taken in the CARESSES project, its initial steps, and its future plans.
\end{abstract}

% EU-JP H2020 project (Culture-Aware Robots and Environmental Sensor
% Systems for Elderly Support, www.caressesrobot.org) 

\newpage
%======================================================================
%======================================================================
\section{Introduction}
\label{sec:introduction}
%======================================================================

%\textbf{** 1 page **}

%\textbf{Why is cultural competence important in assistive robot?}

%\textit{Refer to section 1.1, pages 3-4 of the proposal, without the list of objectives}

%paragraph 1 keep 

Designers of personal assistive robots are often faced with questions
such as: ``How should the robot greet a person?'', ``Should the robot
avoid or encourage physical contact?'', ``Is there any area of the house
that it should consider off-limits?''. Intuitively, the correct answer
to all those questions is ``It depends'', and more precisely, it depends
on the person's values, beliefs, customs and lifestyle.  In one word, it
depends on the person's own \textit{cultural identity}.

%paragraph 2 keep 

The need for cultural competence in healthcare has been widely
investigated in the nursing literature \cite{Leininger02}. The fields of
Transcultural Nursing and Cultural Competence play a crucial role in
providing culturally appropriate nursing care, as the presence of
dedicated cultural competence international journals and worldwide
associations reflects \cite{ETNA_online,CARE_online}. 

%paragraph 2 keep 

In spite of its crucial importance, cultural competence has been almost
totally neglected by researchers and developers in the area of assistive
robotics.  Today it is technically conceivable to build robots ---
possibly operating within a smart ICT environment
\cite{Wongpatikaseree12} --- that reliably accomplish basic assistive
services. However, state-of-the-art robots consider only the problem of
``what to do'' in order to provide a service: they produce rigid
recipes, which are invariant with respect to the place, person and
culture.  We argue that reasoning only about ``what to do'' is not
sufficient and necessarily doomed to fail: if service robots are to be
accepted in the real world by real people, they must take into account
the cultural identity of the assisted person in deciding ``how'' to
provide their services.

\begin{figure}[p]
\centering
\includegraphics[width=\columnwidth]{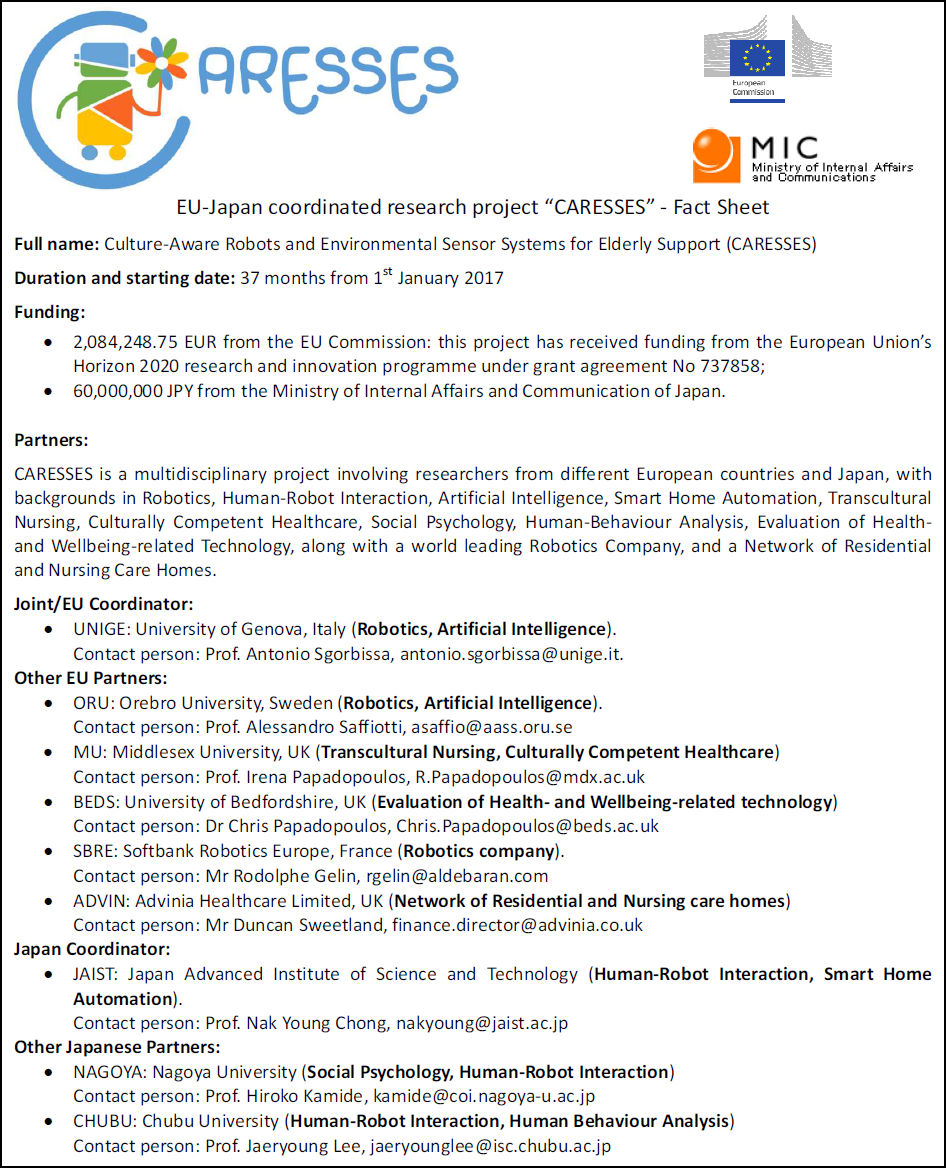}
\caption{CARESSES Fact Sheet.}
\label{fig:fact}
\end{figure}

The CARESSES project%
\footnote{Culture-Aware Robots and Environmental Sensor Systems for
  Elderly Support, www.caressesrobot.org}
is a joint EU-Japan effort that will design culturally aware and
culturally competent elder care robots --- see the Fact Sheet in
Figure~\ref{fig:fact} at the end of this paper.  These robots will be
able to adapt how they behave and speak to the culture, customs and
manners of the person they assist.  The CARESSES innovative approach
will translate into care robots that are designed to be sensitive to the
culture-specific needs and preferences of elderly clients, while
offering them a safe, reliable and intuitive system, specifically
designed to support active and healthy ageing and reduce caregiver
burden.
From a commercial perspective, the cultural customization enabled by
CARESSES will be crucial in overcoming the barriers to marketing robots
across different countries.

The rest of this article describes the CARESSES project in some details.
We start by clarifying, in Section~\ref{sec:facets}, the type of
``cultural competence'' that we address.  Section~\ref{sec:caresses}
gives an overview of the CARESSES approach.  The following three
sections present the developmental methodology, the technical solutions,
and the evaluation strategies, respectively.  Finally,
Section~\ref{sec:status} discusses the project's status and future
plans, and Section~\ref{sec:conclusions} concludes.

% Section \ref{sec:problem_statement} defines the concept of
% \textit{culturally competent robot} and details the capabilities
% enabling culturally competent robot behaviours; Section
% \ref{sec:background} revises the state-of-the-art in such key
% capabilities and Section \ref{sec:methodology} describes CARESSES'
% proposal for a culturally competent robot. Conclusions follow. 

%======================================================================
% EOF
%======================================================================

%======================================================================
\section{Facets of cultural competence}
\label{sec:facets}
%======================================================================

Culture is a notoriously difficult term to define.  In the CARESSES
project, we have adopted the following definitions~\cite{Papadopoulos06}
in order to make our discussion precise and to help us to identify the
key components of a culturally competent robot.

\textbf{Culture}. All human beings are cultural beings. Culture is the
shared way of life of a group of people that includes beliefs, values,
ideas, language, communication, norms and visibly expressed forms such
as customs, art, music, clothing, food, and etiquette. Culture
influences individuals' lifestyles, personal identity and their
relationship with others both within and outside their culture. Cultures
are dynamic and ever changing as individuals are influenced by, and
influence their culture, by different degrees. 

\textbf{Cultural identity}. The concept of identity refers to an image
with which one associates and projects oneself. Cultural identity is
important for people's sense of self and how they  relate to
others. When a nation has a cultural identity it does not mean that it
is uniform. Identifying with a particular culture gives people feelings
of belonging and security. 

\textbf{Cultural awareness}. Cultural awareness is the degree of
awareness we have about our own cultural background and cultural
identity. This helps us to understand the importance of our cultural
heritage and that of others, and makes us appreciate the dangers of
ethnocentricity. Cultural awareness is the first step to developing
cultural competence and must therefore be supplemented by cultural
knowledge. 

\textbf{Cultural knowledge}. Meaningful contact with people from
different ethnic groups can enhance knowledge around their health
beliefs and behaviours as well as raise understanding around the
problems they face. 

\textbf{Cultural sensitivity}. Cultural sensitivity entails the crucial
development of appropriate interpersonal relationships. Relationships
involve trust, acceptance, compassion and respect as well as
facilitation and negotiation. 

\textbf{Cultural competence}. Cultural competence is the capacity to
provide effective care taking into consideration people's cultural
beliefs, behaviours and needs. It is the result of knowledge and skills
which we acquire during our personal and professional lives and to which
we are constantly adding. The achievement of cultural competence
requires the synthesis of previously gained awareness, knowledge and
sensitivity, and its application in the assessment of clients' needs,
clinical diagnosis and other caring skills. 

An analysis of the literature on personal, social and assistive robots
reveals that the issue of culture competence has been largely
under-addressed, and that a lot of work is still to be done to pave the
way to culturally competent robot.

Several studies support the hypothesis that people from different
cultures not only (i) have different preferences concerning how the
robot should be and behave \cite{Evers08}, but also (ii) tend to prefer
robots better complying with the social norms of their own culture
\cite{Trovato15}, both in the verbal \cite{Wang10,Andrist15} and
non-verbal behaviour \cite{Eresha13,Joosse14b}.
Despite these findings, little work has been reported on how to build
robots that can be easily adapted to a given cultural identity.
An interesting example is the framework for the learning and selection
of culturally appropriate greeting gestures and words proposed by
Trovato et al \cite{Trovato15b}.
This work, like all the ones mentioned above, consider adaptation at a
personal level, and follow a ``bottom-up'' approach, i.e., they identify
nations as clusters of people with similar cultural profiles.  As such,
adaptation to a different culture is a demanding process which requires
either a long time, or a large corpus of data to begin with.

A ``top-down'' approach would be better suited for encoding cultural
information expressed at national-level, and how such information
influences preferences in the robot behaviours.
Among the most popular metrics for a top-down description of culture at
a national-level, Hofstede's dimensions for the cultural categorization
of countries are six scales in which the relative positions of different
countries are expressed as a score from 0 to 100 \cite{Hofstede91}. As
an example, the dimension of \textit{Individualism} examines whether a
nation has a preference for a loosely-knit social framework, in which
individuals are expected to take care of only themselves and their
immediate families, or for a tightly-knit framework, in which
individuals can expect their relatives or members of a particular
in-group to look after them, a notion which Hofstede called
\textit{Collectivism}.
Hofstede's dimensions have been used in the field of assistive robotics
to express the influence of culture on the gestures and words that a
robot should use at a first meeting with a person \cite{Lugrin15}, or to
decide certain motion parameters like the approach distance depending on
the cultural profile of the person \cite{Bruno17}.

%======================================================================
% EOF
%======================================================================

%======================================================================
\section{The CARESSES approach}
\label{sec:caresses}
%======================================================================

The CARESSES approach to design \emph{culturally competent robots}
combines the above top-down and bottom-up approaches.  When a robot
interacts with a person for the first time, it uses a top-down approach
to bootstrap its behavior using a cultural identity based on his or her
cultural group; over the course of time, the robot then uses a bottom-up
approach to refine this cultural identity based on the individual
preferences expressed by that person.

\begin{figure}[tb]
\centering
\includegraphics[width=0.9\columnwidth]{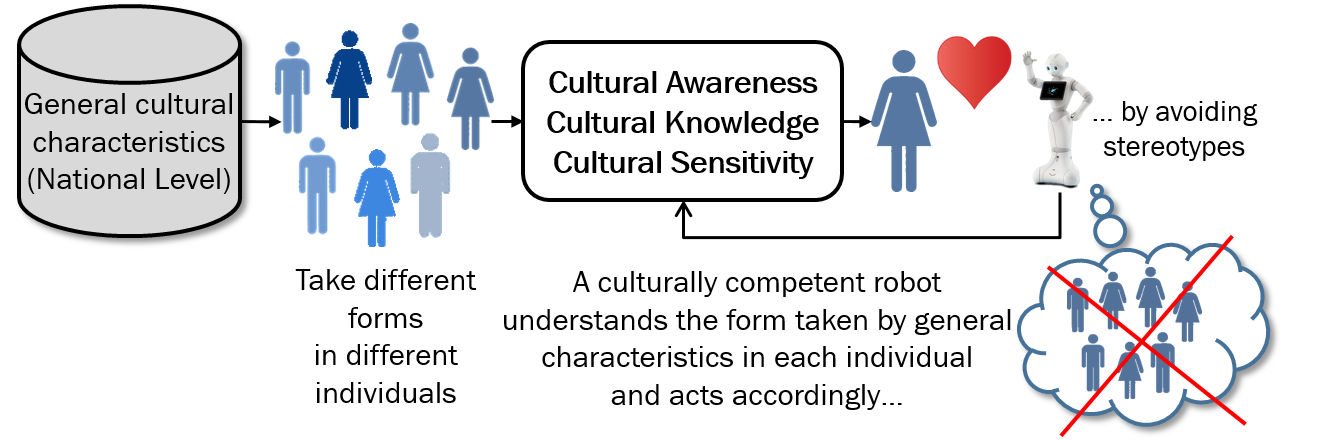}
\caption{The CARESSES concept of a culturally competent robot.}
\label{fig:idea}
\end{figure}

Figure \ref{fig:idea} illustrates this concept.  For CARESSES, a
culturally competent robot: (i) knows general cultural characteristics,
intuitively, characteristics that are shared by a group of people; (ii)
is aware that general characteristics take different forms in different
individuals, thus avoiding stereotypes; and (iii) is sensitive to
cultural differences while perceiving, reasoning, and acting.

More concretely, the culturally aware solutions developed in CARESSES
will be used to expand the capabilities of the Pepper robot, which is
designed and marketed by Softbank Robotics, a partner of the project.
The new culturally aware capabilities will include:

\begin{itemize}
\item communicating through speech and gestures;
\item moving independently;
\item assisting the person in performing everyday tasks, e.g., helping
  with to-do lists and keeping track of bills, suggesting menu plans; 
\item providing health-related assistance, e.g. reminding the person to
  take her medication; 
\item providing easy access to technology, e.g., internet, video calls,
  smart appliances for home automation; 
\item providing entertainment, e.g., reading aloud, playing music and
  games.
\end{itemize}

One of the key questions asked in CARESSES is what added value does
cultural competence brings to an assistive robots.  In order to
precisely answer this questions, the CARESSES culturally competent
robots will be systematically evaluated at different test sites in
Europe and in Japan, namely: the Advinia Healthcare care homes (UK;
project partner); the HISUISUI care home (Japan); and the iHouse
facility at JAIST (Japan; project partner).  These facilities play
complementary roles in the evaluation: the Advinia and HISUISUI care
homes provide access to real end users, who will take part in the
evaluation; while the iHouse, a duplex apartment fully equipped with
sensors and smart appliances for home automation, will allow us to
explore the integration of culturally competent robots in smart
environments.

\begin{figure}[tb]
\centering
\includegraphics[width=0.9\columnwidth]{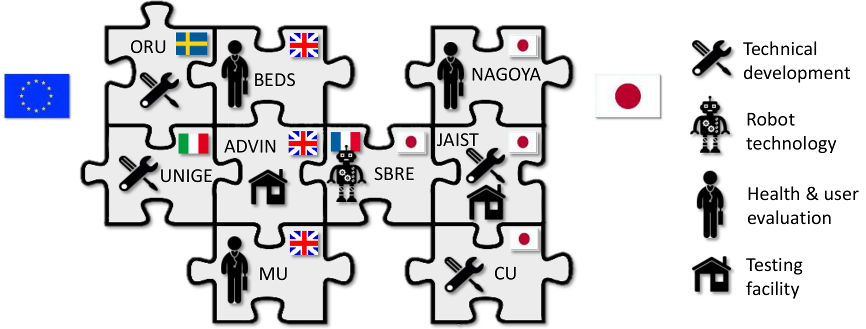}
\caption{The CARESSES international consortium.  See the Fact Sheet in
  Figure~\ref{fig:fact} for a list of partners and their short names.}
\label{fig:consortium}
\end{figure}

The ambitious role of CARESSES require the close interaction of actor
from different areas and different sectors.  Accordingly, the CARESSES
consortium includes experts in trans-cultural nursing, in robotics, in
artificial intelligence, in human-robot interaction, and in the
evaluation of technologies for the elderly.  These are complemented by
end users, and by a company with a long experience in developing and
marketing assistive, social, educational and entertainment robots.
Figure~\ref{fig:consortium} shows the competences and geographical
distribution of the partners in the CARESSES consortium.

Cooperation between the EU and Japan is key for the successful
accomplishment of the objectives of CARESSES.  First, it offers the
possibility to investigate the outcomes of interest of CARESSES through
a robust user-testing approach: the impact of cultural competence in
assistive robotics will be evaluated in very different cultural contexts
in EU and Japan.  Second, it allows CARESSES to benefit from the
complementary competences and resources of EU and JP partners.  Among
the others, EU and JP Participants will pool two extremely valuable
resources for system integration and testing: (i) the universAAL
software platform for open distributed systems of systems
\cite{Ferro15}; and (ii) the iHouse facility at JAIST.  Finally, this
cooperation is expected to foster Transcultural Studies in Japan.  While
the study of Transcultural Nursing is supported by international
journals and associations in the EU \cite{ETNA_online}, similar studies
or associations do not yet exist in Japan.  As the number of foreign
residents in Japan is significantly increasing, CARESSES will contribute
to draw the attention of health-care professionals and public
decision-makers to the role of cultural competence in health-care.

The project management structure has been designed to adequately address
the peculiarities of CARESSES, a joint EU-JP project whose research
activities are performed in the context of two coordinated projects
funded by different agencies: the EU Commission, and the Japanese
Ministry of Internal Affairs and Communications.
The management of CARESSES also benefits from the support of an Ethical
Advisor, of an Exploitation, Dissemination, and IPR Board, and of an
External Expert Advisory Board.

%======================================================================

%======================================================================
\section{Development methodology}
\label{sec:methodology}
%======================================================================

CARESSES follows a strongly user-driven methodology to develop relevant
technological solutions, where requirements and knowledge come from the
end users and domain experts, and validation is done on the end users.

The research in the areas of Transcultural Nursing and Culturally
Competent Healthcare constitutes the theoretical foundations of
CARESSES, paving the way to the development of culturally competent
robots.
The robot's attitude towards users has been initially created based on
data available on the Hofstede's dimensions \cite{Hofstede91},
complemented with specific information about the user's cultural group.
This has allowed making preliminary assumptions about the expected
behaviour of a cultural competent robot, described in terms of Cultural
Awareness, Cultural Knowledge and Cultural Sensitivity
\cite{Papadopoulos06}.

%----------------------------------------------------------------------
% Table I
%----------------------------------------------------------------------

\begin{table*}[t]
\caption{Introduction scenario: Mrs Christou, a 75 years old Greek
  Cypriot who migrated to the UK when she was 20 years old.} 
\label{tab:Introduction}
\begin{center}
\resizebox{\columnwidth}{!}{
\begin{tabular}{|p{6.8cm}|p{3.8cm}|p{5.9cm}|}
\hline
Scenario & Robot skills & Cultural competence \\
\hline
ROBOT: Hello Mrs. Christou! & Perception (Face recognition) & \\
\textit{The robot hugs Mrs. Christou} & Moving (Arms) & \\
MRS CHRISTOU: Hello! & & \\
\textit{Mrs. Christou smiles and hugs the robot} & & \\
ROBOT: Would you prefer me to call you Kyria Maria? & Speaking (Asking for yes/no confirmation) & [Cultural Knowledge: general (1)] The Greek Cypriot culture is very similar to that of Greece, in which hierarchy should be respected and some inequalities are to be expected and accepted.\\
MRS CHRISTOU: Yes, that's how one calls an older woman in Cyprus. What is your name? &  & [Cultural Awareness (2)] Mrs. Christou values her culture and its customs. She expects others to treat her older age status with some respect: this is why she likes that the robot calls her Kyria Maria (Kyria is Greek for Mrs).\\
ROBOT: I don't have a name yet. Would you like to give me a name? & Speaking (Catching key words and reacting) & \\
\textit{The robot leans slightly forward} & Moving (Body posture) & \\
KYRIA MARIA: I will call you Sofia after my mother, God rest her soul. & & [Cultural Awareness (3)] She names the robot after her mother, a common custom to name one's children. She shows her respect to the dead through signs of her religiosity. \\
\textit{The robot asks for confirmation for the name, infers that Sofia is the name of Kyria Maria's mother and asks for confirmation} & Speaking (Catching key words; asking for yes/no confirmation) & \\
ROBOT SOFIA: Thank you, I like the name. I am honoured to be called after your mother. & &\\
\textit{The robot smiles and hugs Kyria Maria} & Moving (Arms) & \\
\hline
\end{tabular}}
\end{center}
\end{table*}

%----------------------------------------------------------------------
% Table II
%----------------------------------------------------------------------

\begin{table*}[t]
\caption{Health-care scenario: Mrs Smith, a 75 year old English lady, a
  former school teacher.} 
\label{tab:Healthcare}
\begin{center}
\resizebox{\columnwidth}{!}{
\begin{tabular}{|p{6.8cm}|p{3.8cm}|p{5.9cm}|}
\hline
Scenario & Robot skills & Cultural competence\\
\hline
\textit{The robot Aristotle detects that Mrs. Smith is in a bad mood and adopts a more cheerful voice} & Perception (Understanding facial expressions) & \\
ROBOT ARISTOTLE: How do you feel today Dorothy? & &\\
MRS DOROTHY SMITH: I feel OK but it's time for my tablets. I have diabetes. & & [Cultural Knowledge: general (4)] The UK has a pragmatic orientation.\\
A: Do you take tablets for diabetes? & Speaking (Catching key words; asking for yes/no confirmation) & [Cultural Knowledge: specific (5)] The robot is matching what Mrs. Smith says with pre-stored knowledge about her health.\\
D: Yes. &  & \\
A: Do you want me to remind you to take them? & &\\
D: Yes! I take them three times a day: morning, midday and evening. But sometimes I forget them. & &\\
A: OK. I will remind you! Please select your schedule on my screen. & Planning (Reminder) & [Cultural Knowledge: specific (6)] The robot knows that Mrs. Smith, a former school teacher, is already familiar with using a tablet. \\
\textit{The robot leans forward. Mrs. Smith selects morning, midday and evening on the screen} & Moving (Body posture), Multi-modal Interaction (Using multiple input modalities) & \\
A: Is there anything I can do for you? Can I get you some water for the tablets? & & [Cultural Knowledge: specific (7)] The robot is acquiring knowledge about what it means to Mrs. Smith to have diabetes.\\
D: Yes. That would be very nice Aristotle. & &\\ 
\textit{The robot goes to fetch a glass of water} & Planning (Retrieving an object), Perception (Locating an object), Moving (Legs, hands) &\\
\hline
\end{tabular}}
\end{center}
\end{table*}

%----------------------------------------------------------------------
% Table III
%----------------------------------------------------------------------

\begin{table*}[t]
\caption{Home and family scenario: Mrs Yamada, a 75 years old Japanese lady who performed tea ceremony in Kobe for 40 years}
\label{tab:Home_and_family}
\begin{center}
\resizebox{\columnwidth}{!}{
\begin{tabular}{|p{6.8cm}|p{3.8cm}|p{5.9cm}|}
\hline
Scenario & Robot skills & Cultural competence\\
\hline
ROBOT YUKO: It is possible to test the video call with your family, if you like it. & Speaking (Avoiding direct questions) &\\
\textit{The robot checks for Mrs. Yamada's reaction. She smiles.} & Perception (Understanding facial expressions) &\\
MRS NAOMI YAMADA: Really? My son and daughter both live in Tokyo. My son is always busy, but he visits me during holidays. I miss my daughter so much. Her husband is Korean so she often goes to Korea. I want to call my husband, but he's now giving a lecture at school. & & [Cultural Knowledge: specific (8)] Naomi provides her personal details only when the robot brings up the topic.\\
Y: I can make a video call to your daughter, son or husband if you want. & Speaking (Catching key words and reacting) &\\
\textit{The robot checks for Mrs. Yamada's reaction} & Perception (Understanding facial expressions) &\\
N: Maybe later. I don't know how to do it. Can you give me a manual on how to do it? & & [Cultural Knowledge: general (9)] Japan is one of the most uncertainty avoiding countries on earth.\\
Y: Just tell me who you want to call, then I can help you. You are welcome to try. & Planning (Video call) & [Cultural Sensitivity (10)] Empowering: the robot is sensitive of the fact that Naomi is frightened by using unknown technology, and encourages her.\\
N: Ok, let's try. You will be my assistant! & &\\
\hline
\end{tabular}}
\end{center}
\end{table*}

%----------------------------------------------------------------------
\subsection{Initial scenarios}
%----------------------------------------------------------------------

In order to arrive to a set of technical requirement, we have grounded
the above concepts in concrete examples.  These are summarized in Tables
\ref{tab:Introduction}, \ref{tab:Healthcare} and
\ref{tab:Home_and_family}, that describe possible scenarios of
interaction between a culturally competent assistive robot and an
elderly person.  The scenarios have been written by experts in
Transcultural Nursing and draw inspiration from the rationale and
actions of culturally competent caregivers.  Each table reports a
pattern of sensorimotor and/or verbal interaction, the required robot
skills, as well as the cultural competence (in terms of cultural
awareness, cultural knowledge and cultural sensitivity) that may
contribute to determine the robot's behaviour.  

Albeit short, these scenarios show that the following capabilities are
key for a robot to exhibit a culturally competent behaviour:

\begin{itemize}
\item \textit{cultural knowledge representation}, this refers to the
  capability of storing and reasoning upon cultural knowledge, see for
  example the interaction between the robot Aristotle and Mrs. Smith in
  Table \ref{tab:Healthcare}, in which the robot first uses knowledge
  (6) about Mrs. Smith 's work experience to tune how to introduce a new
  interaction modality (its tablet), and later acquires new knowledge
  about her habits and medical prescriptions (7);

\item \textit{culturally-sensitive planning and execution}, this refers
  to the capability to produce plans and adapt such plans depending on
  the cultural identity of the user. Cultural sensitivity, in the
  interaction between the robot Yuko and Mrs. Yamada in Table
  \ref{tab:Home_and_family}, allows the robot for planning to help
  Mrs. Yamada make a video call (10);

\item \textit{culture-aware multi-modal human-robot interaction}, this
  refers to the capability of adapting the way of interacting (in terms
  of gestures, tone and volume of voice, etc.) to the user's cultural
  identity. Cultural sensitivity makes the robot avoid asking direct
  questions to Mrs. Yamada (see Table \ref{tab:Home_and_family}) and
  perform the proper greeting gestures with Mrs. Christou (see Table
  \ref{tab:Introduction});

\item \textit{culture-aware human emotion and action recognition}, this
  refers to the capability to interpret sensor data acquired by the
  robot during the interaction in light of cultural knowledge. As an
  example, in Table \ref{tab:Healthcare} the robot Aristotle correctly
  labels Mrs. Smith facial expression as indicative of a bad mood, while
  in Table \ref{tab:Home_and_family} the robot Yuko relies on
  Mrs. Yamada's facial expression to get feedback on its suggestion to
  make a video call; 

\item \textit{cultural identity assessment, habits and preferences
  detection}, this refers to the capability to adapt general cultural
  knowledge and acquire new knowledge to better fit the individual
  profile of the user. As an example, in Table \ref{tab:Introduction}
  the robot Sofia uses knowledge about the Greek culture to guess how
  Mrs. Christou would like to be addressed (1), and her answer to
  validate its hypothesis (2). In Table \ref{tab:Healthcare}, the robot
  Aristotle learns Mrs. Smith's habits in dealing with her medical
  prescriptions (5), and in Table \ref{tab:Home_and_family} the robot
  Yuko brings up the topic of video calls (8) to learn about
  Mrs. Yamada's family.

\end{itemize}

%----------------------------------------------------------------------
\subsection{Elicitation of guidelines}
%----------------------------------------------------------------------

To ground the assumptions made into observations in real-world
scenarios, the robot's cultural competence is now undergoing a process
of iterative refinement on the basis of the cultural behavioural cues
collected from video recorded encounters between older people living in
sheltered housing and their caregivers. Specifically, having identified
and verified the relevant verbal and non-verbal behavioural cues, a
panel of experts shall then refine the assumption made and update the
prototype robot's cultural competence. In doing this, a great care
should be paid to eliminate stereotypic notions \cite{Makatchev13}.

This process will ultimately produce \emph{guidelines} describing how
culturally competent robots are expected to behave in assistive
scenarios. The knowledge acquired in all these steps, both through
comprehensive literature reviews on the topics and video recorded
encounters, shall be formalized using tools for knowledge
representations, as the availability of formal languages for knowledge
representation constitutes the basis for the robot to exhibit autonomous
reasoning, planning and acting skills depending on such knowledge. Also,
in the perspective of a commercial exploitation, it will allow the
development of robots that are able to autonomously acquire information
and update their own knowledge about the cultural context in which they
are operating and – ultimately – to re-configure their approach towards
the user.

%======================================================================
% EOF
%======================================================================

%====================================================-*-Mode: LaTeX -*-
\section{Technical solutions}
\label{sec:technology}
%======================================================================

From a technological point of view, the first steps of CARESSES have
been to: (i) identify the basic technical capabilities required for
cultural competence, and (ii) define a system architecture that enables
the integration of these capabilities into a fully autonomous assistive
robot, possibly embedded in a smart environment.

%----------------------------------------------------------------------
\subsection{System architecture}
%----------------------------------------------------------------------

\begin{figure}[tb]
\centering
\includegraphics[width=\columnwidth]{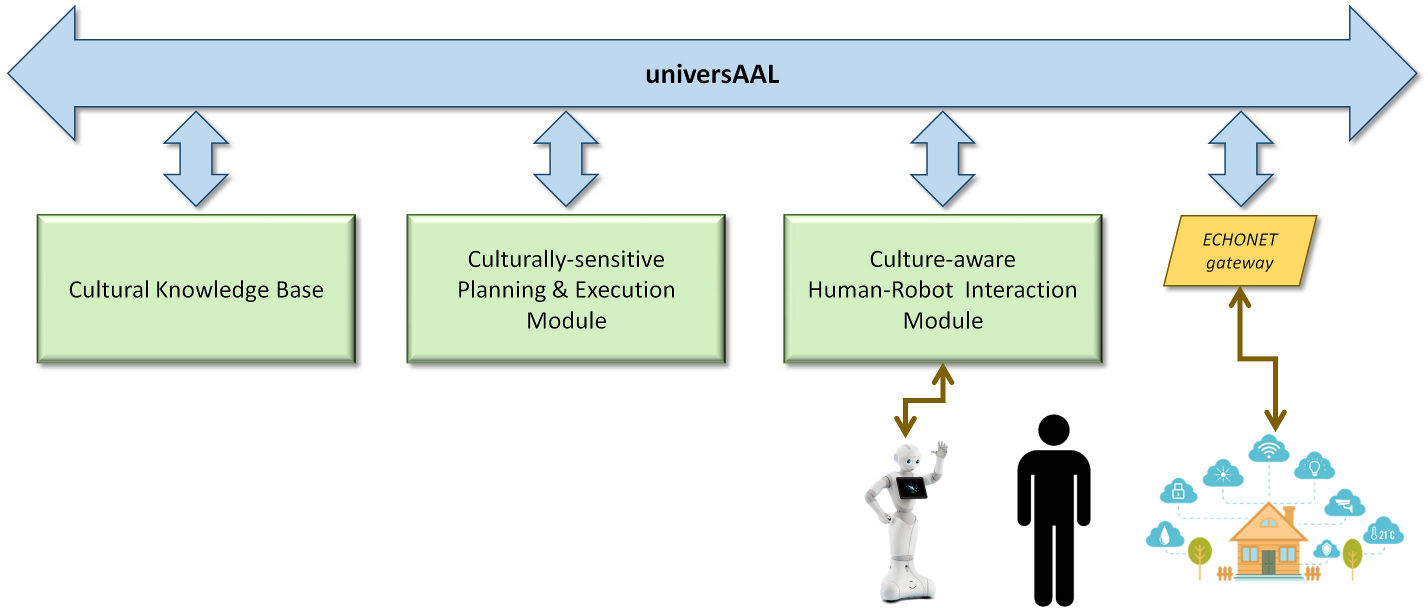}
\caption{A coarse view of the CARESSES architecture.}
\label{fig:architecture}
\end{figure}

Figure~\ref{fig:architecture} shows the main modules of the CARESSES
architecture: the \textit{Cultural Knowledge Base}, the
\textit{Culturally-sensitive Planning \& Execution}, and the
\textit{Culture-aware human-robot interaction}, which are briefly
described in the next sub-sections.

% For sake of brevity, we do not provide details about the semantics of
% the exchanged message, whose meaning is however straightforward in most
% cases.

All the components are being integrated in universAAL~\cite{Ferro15}, a
software platform for open distributed systems of systems that resulted
from a consolidation process conducted within an EU Project.  The
universAAL platform allows the seamless integration of heterogeneous
components within an environmental network through the extensive
exploitation of two base concepts, that turn out to be very suited for
integration of components developed by different partners of a complex
RTD project: (i) the usage of ontologies to define the services provided
by network nodes as well as the procedures and data formats to access
such services; (ii) the usage of three dedicated communication buses for
topic-based communication among network nodes, namely a Context Bus, a
Service Bus, and a User Interface Bus.

A bridge between universAAL and ECHONET~\cite{Matsumoto10}, the Japanese
ECHONET standard for home automation, is being designed.  This will
allow the robot to communicate with a smart environment.  In CARESSES,
this will be provided by the iHouse, a Japanese-based duplex apartment
fully embedded with sensors and actuators for home automation.

%----------------------------------------------------------------------
\subsection{Cultural Knowledge Base}
%----------------------------------------------------------------------

Properly encoding guidelines for cultural competence in a framework for
knowledge representation requires to take into account both
methodological and architectural aspects.

Methodological aspects include: (i) how to represent the relationship
between quantitative and qualitative knowledge about different cultural
groups; (ii) how to avoid stereotypes by allowing for differences among
individuals, while using the information about their national culture as
a hint about their cultural identity; (iii) how to automatically reason
on cultural knowledge for producing a culturally competent robotic
behaviour, i.e., plans and sensorimotor behaviours aligned with the
user's cultural identity; and (iv) how to update the knowledge base as
long as new cultural knowledge is acquired through user-robot
interaction.

Architectural aspects include: (i) which languages and tools should be
used for knowledge representation, e.g., Description Logics
\cite{Baader03}, Bayesian networks \cite{Pearl00}, OWL probabilistic
extensions such as PR-OWL \cite{Laskey08}; (ii) which languages and tools
should be used for querying the knowledge base: e.g., standard OWL
functions, SPARQL;%
\footnote{https://www.w3.org/TR/rdf-sparql-query/}
(iii) which reasoning tools should be adopted, e.g., basic mechanisms
such as subsumption or instance checking, rule-based languages such as
SWRL,%
\footnote{https://www.w3.org/Submission/SWRL/}
Bayesian inference; and (iv) which Application Programming Interfaces,
data formats, and protocols should be used to allow the robot to access
the knowledge base.

% Both the two aspects above shall be considered, that shall ultimately

CARESSES will consider the above aspects to produce a portfolio of
different solutions to be included in the general framework for
Knowkedge Representation.  It will also define procedures for the
knowledge base creation and updating.  Indeed, Cultural Knowledge can
be: (i) defined and introduced in the system \textit{a priori} by
experts in Transcultural Nursing, formal and informal caregivers; (ii)
acquired at run-time through robot-user interaction.  Specifically,
run-time knowledge acquisition raises the most important methodological
and technological issues, e.g., in terms of which questions should be
posed to the user, how answers should be interpreted, how the
information retrieved should be used to pose subsequent questions and to
update the Knowledge Base itself.  It also raises issues on how general
cultural information known a priori (e.g., at National level) impact on
individual characteristics, and how the information acquired during
robot-user interaction (i.e., observed through sensors or explicit
communication) can be merged with the already available knowledge in
order to perform a more accurate cultural assessment.

Finally, Knowledge Representation raises ethics issues in data privacy
and protection, extremely relevant whenever the system stores sensitive
information not only about the user, but also his / her family (e.g.,
names, domicile).

%----------------------------------------------------------------------
\subsection{Culturally-Sensitive Planning and Execution}
%----------------------------------------------------------------------

Once cultural knowledge has been explicitly produced, the challenge is
to make the robot use this knowledge to modulate its own behaviour to
match the cultural identity of the user. Technically, this translates
into the ability to: (i) form plans to achieve the robot's goals while
being aware of, and sensitive to, the user's culture; and (ii) execute
the actions in these plans in a way that is also culturally aware and
sensitive.  As an example, the three robots in Tables
\ref{tab:Introduction}, \ref{tab:Healthcare} and
\ref{tab:Home_and_family} may have the same goal to help preparing the
lunch, but they may achieve this goal using different plans. These plans
may include different actions (e.g., Aristotle may help Mrs Smith by
ordering the food online, whereas Sofia listens to Mrs Christou chatting
about cooking), or different ways to perform an action (e.g., Yuko
collaboratively prepares the lunch with Mrs Yamada).

The field of Artificial Intelligence (AI) has a long tradition in
developing techniques for the automatic generation and execution of
action plans that achieve given goals \cite{Ghallab14}. Cultural aspects
can contextually influence the generation and execution of action plans
in three ways:
\begin{itemize}
\item Discourage the use of certain actions; for example, to avoid
  suggesting recipes to Mrs Christou; 
\item Include additional preconditions or goals, which may result in the
  inclusion of new actions; for example, with Mrs Yamada, the robot Yuko
  performs an inquiry action before committing to one action plan or
  another; 
\item Induce a preference for some actions; for example, Yuko may
  encourage Mrs Yamada to cook instead of ordering food online, because
  this better complies with Mrs Yamada's need to make physical
  activity. 
\end{itemize}

To take these influences into account, state-of-the-art approaches to
constraint-based planning \cite{Mansouri16} have been considered. In
addition to requirements in terms of causal preconditions (e.g., the
robot's hand must be empty to grasp an object), spatial requirements
(e.g., the robot must be in front of the user in order to interact), and
temporal constraints (e.g., the tea must be served before it gets cold),
constraint-based planning can also include constraints that pertain to
the human-robot relation, e.g., to encode the fact that the robot should
never clean a room where the user is standing: this extension of
constraint-based planning is particularly suited to generate plans that
take into account cultural constraints and, in general, ``human-aware
planning'' \cite{Kockemann14}. 

%----------------------------------------------------------------------
\subsection{Culture-aware Human-Robot Interaction}
%----------------------------------------------------------------------

Once a proper course of actions (including both motion and speech) has
been planned taking into account the user's cultural identity, actions
must be executed and feedback must be considered to monitor their
execution. In this context, Human-Robot Interaction plays a crucial role
in enabling the robot with cultural competence. On the one hand, the way
the robot behaves and speaks can produce different impacts and
subjective experiences on the user; on the other hand, what the user
says and does is the key for the robot to acquire new knowledge about
the user, and consequently refine and improve its cultural competence.

% Indeed, as already discussed, the robot must be able to perform a
% cultural assessment of the user through explicit communication and
% through observation: this is the key for interacting properly
% depending on the user’s individual cultural characteristics in the
% light of general cultural characteristics. 

As a prerequisite, the robot shall be equipped with motor capabilities
that are sophisticated enough to allow it to exhibit its cultural
competence through motions, gestures, posture, speech; similarly, it is
mandatory that the robot (and possibly  the  environment) is equipped
with sensors and devices for multimodal audio / video / haptic
interaction that allow providing feedback to the modules for planning,
action execution and monitoring, as well as perceiving the nuances of
human behaviour in different cultures. Moreover, communication devices
allowing for a simplified interaction may also be fundamental for frail
older adults. 

The role played by robot-user \textit{verbal communication} has been
carefully considered, as it is the primary way of interaction, possibly
allowing to acquire new knowledge and update the Cultural Knowledge
Base. Due to the current limitation in natural language understanding,
semantic comprehension is limited to the recognition of relevant
keywords, that the robot will use to react accordingly, by asking a
confirmation through a simple multiple choice (e.g., yes/no) question.
Additional touchscreen-based interfaces (either embedded on the robot or
carried by users, e.g., tablets and smartphones) are used to complement
the verbal interaction modality.

The robot's perceptual capabilities  include the ability to estimate
human \textit{emotions} (joy, sadness, anger, surprise) and recognize
human \textit{actions}. As the robot operates in a smart ICT
environment, the usage of lightweight wearable sensors that do not
interfere with daily activities is explored (e.g., smartwatches or
sewable sensors).

% Emotions and actions will constitute the basis for to assessing the
% cultural identity of individuals by observing their daily lives. 

Lastly, the robot will be equipped with a module to detect and recognize
\textit{daily activities}, i.e., combinations of primitive actions
performed in different contexts and places of the house (e.g., walking,
cleaning, sitting on a sofa, etc.) \cite{Bruno14}. As time progresses
and the robot has more and more interactions with the user, daily
activities and manners (a subset of social norms that regulate the
actions performed by the user towards other humans, or even the robot
itself) may be assessed to determine the long-term \textit{habits} of
the human companion.  

Verbal interaction, as well as the assessment of user's, emotions,
actions, daily activities, manners, and habits, will ultimately provide
an input to perform a cultural assessment of the user, updating the
knowledge that the robot has about the user's cultural identity.

The aforementioned capabilities involve procedures to merge and
interpret sensor data acquired by the robot and by the smart ICT
environment at the light of cultural knowledge that is already stored in
the system.  Indeed, cultural knowledge can play a fundamental role at
all levels of perception, ranging from basic object recognition to the
detecting of daily activities, manners and habits.  For instance, if the
system is uncertain if a purple object in the fridge is a slice of pig
liver or an eggplant, cultural information about the alimentary customs
of the users (that maybe are vegetarians) could help to disambiguate.

%======================================================================
% EOF
%======================================================================

%======================================================================
\section{Evaluation strategy}
\label{sec:evaluation}
%======================================================================

To the end of testing and evaluation with elderly participants, an
ethically sensitive and detailed protocol that describes the screening,
recruitment, testing and analytical procedures is being produced and
scrutinised by relevant ethics committees. Once ethical approval is
obtained, testing will commence. Testing will involve older adults
belonging to different cultural groups, who possess sufficient cognitive
competence to participate and who are assessed as sufficiently unlikely
to express aggression during the testing period. Nominated key informal
caregivers (e.g., close family members) shall also be recruited.

% It shall follow a rigorous and ethically approved procedure for
% screening and recruiment, informing and getting consent from
% participants, privacy in data collection and storage. 

The impact of cultural competence will be quantitatively evaluated by
dividing participants in experimental and control arms, i.e.,
interacting with robots with and without cultural customization, by
using state-of-the-art quantitative tools to measure the impact of
healthcare interventions.

% (quantitative tools which, unfortunately, are not very common when
% assessing the impact of assistive robotic solutions).

End-user evaluation will be aimed at evaluating the capability of
culturally competent systems to be more sensitive to the user's needs,
customs and lifestyle, thus impacting on the quality of life of users
and their caregivers, reducing caregiver burden, and improving the
system's efficiency and effectiveness.
The evaluation will take place at the Advinia Health Care network of
Residential and Nursing care homes (UK) as well as in the HISUISUI care
facility (Japan).  The evaluation in the UK care homes will include at
least ten clients who primarily identify themselves with the
white-English culture, ten who primarily identify themselves with the
Indian culture, and ten who primarily identify themselves with the
Japanese culture.  Each group will be divided into an experimental and a
control arm.  Each client will adopt a Pepper robot for a total of 18
hours over a period of two weeks.  This shall allow for enough time for
a culturally customized Pepper robot to acquire knowledge about the
individual cultural characteristics of the assisted person and provide
culturally competent interactions and service, which will then be
evaluated through quantitative tools and qualitative interviews.

% at the 0.05 probability threshold. 

% To this end, end-user evaluation shall rely on state-of-the-art
% measurement tools, possibly adapted to cope with the specific scenario
% addressed by the project. 

Quantitative outcomes of interest and measurement tools shall include
(pre and post testing): 

\begin{itemize}
\item Client perception of the robot's cultural competence. Measurement
  tool: Adapted RCTSH Cultural Competence Assessment Tool (CCATool)
  \cite{Papadopoulos04} that shall measure clients' perceptions of the
  robot's cultural awareness, cultural safety, cultural competence and
  cultural incompetence, and shall include items associated with
  dignity, privacy and acceptability.

\item Client and informal caregiver health related quality of
  life. Measurement tool: Short Form (36) Health Survey (SF-36)
  \cite{Hays93}. The SF-36v2 is a multi-purpose, short-form health
  survey proven to be useful in surveys of general and specific
  populations, including older adults. It measures: general health,
  bodily pain, emotional role limitation, physical role limitation,
  mental health, vitality, physical functioning and social
  functioning. Each dimension score has values between 0 and 100, in
  which 0 means dead and 100 perfect health.

\item Informal caregiver burden. Measurement tool: The Zarit Burden
  Inventory (ZBI) \cite{Zarit80} is a widely used 22-item self-report
  inventory that measures subjective care burden among informal
  caregivers. Its validity and reliability has been widely
  established. The scale items examine burden associated with
  functional/behavioural impairments and care situations. Each item is
  scored on a 5-point Likert Scale, with higher scores indicting higher
  care burden among informal caregivers.

\item Client satisfaction with the robot. Measurement tool:
  Questionnaire for User Interface Satisfaction (QUIS)
  \cite{Chin88}. This scale evaluates whether the clients are satisfied
  with the interaction process including its efficiency and
  effectiveness. It will be adapted so that ``the software'' is replaced
  by ``the robot''. 

\end{itemize}

Clients and their informal caregivers will also be invited to
participate in qualitative interviews to elicit discussion among their
perceptions of the robot's cultural competence, quality of service
provided, impact upon independence and autonomy. Discussions with
informal caregiver shall focus upon caregiver burden, impact on quality
of life, and -- very importantly -- experiences related to configuring
the system by injecting cultural knowledge before operations.

During evaluation, differences between the pre- and post-testing
measures shall be considered. However, given that this is the first time
that --- to the best of our knowledge --- the above tools are proposed
to be applied in a robotic context, there are no sufficiently similar
previous experimental outcome data to use to statistically power our
trial in order to sensitively detect clinically meaningful differences
in pre- and post- outcome measures.  Therefore, we shall describe
observed statistical differences in outcome measures within and between
arms, and report, using appropriate inferential tests, whether these
differences are statistically significant.

%======================================================================
% EOF
%======================================================================

%======================================================================
\section{Current status and future work}
\label{sec:status}
%======================================================================

The CARESSES project started on January 1, 2017.  Its workplan span 37
months, implementing the methodology discussed above.  The starting
technology includes Softbank's Pepper robot%
\footnote{Pepper is produced by SoftBank Robotics Europe.}
as well as the \textit{iHouse}, a Japanese-based
duplex apartment fully embedded with sensors and actuators for home
automation.%
\footnote{The iHouse has been developed by the Japan Advanced Institute
  of Science and Technology.}
Cultural competence will be achieved through an investigation phase
aimed at producing guidelines for Transcultural Robotic
Nursing%
\footnote{This investigation phase will be led by Middlesex University
  (UK) with the deep involvement of Nagoya University (Japan).} 
as described in Section~\ref{sec:methodology}, and then through the
development of three main technological components described in
Section~\ref{sec:technology} above,%
\footnote{The development of these three components will be led,
  respectively, by University of Genova (Italy), \"{O}rebro University
  (Sweden), and Japan Advanced Institute of Science and Technology, with
  the deep involvement of SoftBank Robotics Europe (France) and Chubu
  University (Japan).}
and integrated in universAAL.  Testing and end-user evaluation will
follow the procedure in Section~\ref{sec:evaluation}.%
\footnote{End-user evaluation will be led by Bedfordshire
  University (UK), with the deep involvement of Nagoya
  University.}

In the first months of the project, all partners of the consortium have
closely cooperated to define the initial specifications, produce
scenarios of human-robot encounters, and design the system's
architecture in all its details.  A skeleton implementation of this
architecture, embedded in the universAAL framework, has been produced.
The design and implementation of the three main modules shown in
Figure~\ref{fig:architecture} has started.

The definition of scenarios is particularly important, as it will drive
the investigation of guidelines for cultural-competence in robotics, the
technical development of the required robotic capabilities (in terms of
motion, perception, planning, knowledge representation, etc.) as well as
the design of experiments for testing and evaluation since the early
stages of the project.  All partners of the consortium, led by Middlesex
University, have cooperated for the production of these scenarios using
online tools for collaborative working (e.g., Google Drive and the
CARESSES gitLab repository).  The close collaboration of domain experts
with the technical partners has been pivotal to the success of this
stage, as it has been the close collaboration between the EU and the
Japanese partners.

The CARESSES project will now proceed through the following major
milestones:

\begin{description}

\item[October 2017] The basic guidelines for the development of
  culturally competent robots are ready to be used to start populating
  the cultural knowlegde base.
  On the technical side, a fully integrated, albeit simplified version
  of the CARESSES system is implemented and integrated in the universAAL
  application layer.  This include a basic version of the cultural
  knowledge base, of the culturally sensitive planning and execution
  module, and of the culture-aware human robot interaction module.  The
  latter includes the main motion and perceptual capabilities required
  by Pepper in the scenarios.  Each component will subsequently be
  iteratively expanded, standalone debugged and tested, and refined up
  to its final form.

\item[October 2018] The detailed protocol that describes the screening,
  recruitment, and testing procedures has been prepared and approved by
  Ethics Committee.
  On the technical side, the culturally competent robot is ready for
  final integration, deployment testing and evaluation. All the software
  modules have been properly integrated in universAAL and standalone
  tested as they were developed, and can be properly configured to
  produce the desired culturally competent behaviour.  The full cultural
  knowledge base has been populated with the guidelines for culturally
  competent robots.

\item[August 2019] Tests in Health-Care Facilities and iHouse completed.
  Data and logs of robot-user encounters have been collected.  These
  include data from pre- and post-testing structured interviews as well
  as Post-testing qualitative semi-structured interviews with clients
  and informal caregivers, ready to be analyzed.

\end{description}

The CARESSES project will end in December 2019.  It will release its
technical components for culturally competent robots as Open Source, in
a form suitable to be executed in the application layer of universAAL.
Most of the CARESSES deliverables are public and they will be widely
disseminated.  The current status of the project its publications can be
accessed at the CARESSES web site \url{www.caressesrobot.org}.

%======================================================================
% EOF
%======================================================================

%====================================================-*-Mode: LaTeX -*-
\section{Conclusions}
\label{sec:conclusions}
%======================================================================

Assistive robots can help foster the independence and autonomy of older
persons in many ways, by reducing the days spent in care institutions
and prolonging time spent living in their own home.  Cultural competence
allows assistive robots to be more acceptable by being more sensitive to
the user's needs, customs and lifestyle.
% 
% thus producing a greater impact on the quality of life of users and
% their caregivers, reducing caregiver burden, and improving the
% system's effectiveness. 
% 
In this article, we have discussed the notion of cultural competece that
is put forward in the context of the CARESSES joint EU-Japan project.
% In CARESSES, a culturally competent robot is able to autonomously
% re-configure its way of acting and speaking, when offering a service, to
% match the culture, customs, and etiquette of the person it is assisting.
We have provided an overview of the project, discussed the problems that
it has to address and its methodology, and presented its initial steps.

To the best of our knowledge, CARESSES is the first attempt to build
culturally competent robots.  We believe that cultural competence is a
necessary, althought so far understudied ingredient for any social,
personal or assistive robot.  In this respect, we are persuaded that
CARESSES is a ground breaking project.  Even if CARESSES only produces a
prototype that shall need to be further evaluated and refined before
drawing definitive conclusions about the impact of cultural competence
in assistive robotics, we claim that our pilot will be invaluable in
paving the way for future similar studies.

%======================================================================
% EOF
%======================================================================

%======================================================================
\section*{Acknowledgements}

The CARESSES project is supported by the European Commission Horizon2020
Research and Innovation Programme under grant agreement No.~737858, and
by the Ministry of Internal Affairs and Communication of Japan.

%======================================================================
\bibliographystyle{plain}
\bibliography{biblio}

%======================================================================
%======================================================================
% \section*{Appendix}
% \label{sec:appendix}
%======================================================================

%======================================================================
% EOF
%======================================================================

\end{document}